\pdfoutput=1

\documentclass[11pt,letterpaper]{article}
\usepackage{naaclhlt2013}
\usepackage{times}
\usepackage{latexsym}
\usepackage{color}
\usepackage{multirow}
\usepackage{graphicx}

\title{Is getting the right answer just about choosing the right words?  The role of syntactically-informed features in short answer scoring.}

\author{Derrick Higgins\textsuperscript{*}, 
  Chris Brew\textsuperscript{\dag},  
  Michael Heilman\textsuperscript{*},  
  Ramon Ziai\textsuperscript{\ddag},  
  Lei Chen\textsuperscript{*},  
  Aoife Cahill\textsuperscript{*},  
  Michael Flor\textsuperscript{*},  \\
  \textbf{Nitin Madnani\textsuperscript{*}},  
  \textbf{Joel Tetreault\textsuperscript{\S}},  
  \textbf{Daniel Blanchard\textsuperscript{*}},  
  \textbf{Diane Napolitano\textsuperscript{*}},  
  \textbf{Chong Min Lee\textsuperscript{*}},  \textbf{and} 
  \textbf{John Blackmore\textsuperscript{*}}  \\
  $\ast$Educational Testing Service\\
  \dag{}Nuance Communications \\
  \ddag{}Tuebingen University \\
  \S{}Yahoo! Labs
}

\date{}

\begin{document}
\maketitle
\begin{abstract}
Developments in the educational landscape have spurred greater
interest in the problem of automatically scoring short answer
questions.  A recent shared task on this topic revealed a fundamental
divide in the modeling approaches that have been applied to this
problem, with the best-performing systems split between those that
employ a knowledge engineering approach and those that almost solely
leverage lexical information (as opposed to higher-level syntactic
information) in assigning a score to a given response.  This paper
aims to introduce the NLP community to the largest corpus currently
available for short-answer scoring, provide an overview of methods
used in the shared task using this data, and explore the extent to
which more syntactically-informed features can contribute to the short
answer scoring task in a way that avoids the question-specific manual
effort of the knowledge engineering approach.
\end{abstract}

\section{Introduction}
This paper aims to demonstrate that ``higher-level'' linguistic
features that encode information such as syntactic relations, topics
referenced, and response structure can make a contribution to the
accuracy and validity of automated methods of short answer scoring.
Although the results of a recent shared task on short answer scoring
seem to indicate that lexical features alone cannot be improved upon,
a more thorough examination of the performance of models using
different sorts of features tells a different story.  In support of
this goal, we also provide an overview of the ASAP short-answer
scoring competition, which has gone largely unnoticed in the
community of NLP researchers working on educational applications.

%% DCH1
%% \CHB{Enigmatic. I'd like this paragraph to instead say what the story
%%   is, that is, what the paper shows.}
%%
%% \MJH{I agree with Chris.  Maybe something like ``different story: we
%%   find that syntactic features improve performance by X''?}
%%
%% \MJH{it might be good to have a roadmap of the paper here,
%%   particularly since this paper does not quite follow the usual NAACL
%%   template, where a lot of space is devoted to describing a fancy
%%   model.  also, I remember some recent ACL asking for ``position
%%   papers''.  It almost seems like we should pitch this as one of
%%   those.}

Research on using computers to score open-ended student responses has
a long history, dating back to Ellis Page's work on automated scoring
of essays \cite{page66,page68}.  Since the very beginning of this
research field, there has been an awareness that agreement with human
raters is a limited evaluation measure.  Page's work demonstrated that
the length of an essay correlates strongly with human ratings.  Such
superficial measures can sometimes do surprisingly well as predictive
mechanisms, despite the fact that they are only marginally related to
the skills and attributes we aim to measure with a writing task (the
test \emph{construct}).

%% DCH2
%% \MJH{It may be worth noting that many, if not
%%   most, other areas in NLP are also dissatisfied with their evaluation
%%   methodologies.  Can we claim some general contribution, or make a
%%   case that educational problems are interestingly different in some
%%   way from, say, MT or parsing?  I imagine some reviewer thinking at
%%   this point that he or she has already seen a dozen papers
%%   complaining about various metrics for different tasks.}

Given a sample of scored test-taker responses it is possible to
identify many potentially measurable linguistic features that
correlate well with score. Some of these features rely on advanced
natural language processing, but many do not.  Given the redundancy of
information encoded in many of these features, and the difficulty of
reliably measuring features that depend on advanced NLP, it is
tempting to focus attention on superficial features that are easy to
extract, and to hope that the redundancy will allow good prediction.
However, a system that relies on superficial features as proxies for
important underlying attributes will fail when it begins to see
answers in which the measureable surface features are no longer
correlated with the underlying attributes.  Unfortunately, such
answers are exactly what is to be expected when a sophisticated
test-taking community begins to analyse the test in the search for
simple ways to get good scores.  Therefore, it is important to
understand the potential of deeper features even when their predictive
contribution to scoring in a research setting is limited.

The field of automated essay scoring has made advances in the
intervening years, allowing the development of features related to
various aspects of the writing construct, including lexical sophistication,
discourse structure, syntactic variety, and grammatical accuracy (cf. \newcite{landauer03}; \newcite{elliot03}; \newcite{attali06}; \newcite{attali2010}; \newcite{yannakoudakis2011}; \newcite{foltz2011}).  The
addition of these features has not only improved the conceptual basis
of scoring but also improved the accuracy of these systems according to traditional
evaluation metrics.

Other automated scoring tasks have not yet progressed to the same
level of maturity.  In particular, not as much work has been focused
to date on automated scoring of short-answer questions.  Such
questions are distinguished from essays by their brevity (eliciing
responses of only a few words or a few sentences), and by the fact
that they are scored according to response content, rather than
quality of written expression.  The scoring rubrics for short-answer
questions often require specific information (e.g., scientific
principles, trends in a graph, or details from a reading passage) to
be included in a response for it to receive credit.

The task of short-answer scoring has received more
attention recently, however, because short-answer questions are expected
to figure prominently in new, computerized state tests currently under
development with Race To the Top funding from the US Department of
Education.  As proved to be the case for the automated essay scoring
task, results on the recent ASAP short answer scoring task 
(described later in Section ~\ref{sec:asap-description}) have demonstrated that
superficial features (in this case, features related to the use of
particular words in a response) are strongly predictive.  We aim to
re-examine the contribution of different sorts of predictive features
on the same dataset of short-answer tasks on which these results were
achieved, and demonstrate that attention to linguistic structure is
empirically valuable in automated scoring of short answers.

\section{Previous Work}
Like the field of automated essay scoring, research on methods for
automated scoring of short answer questions has a history that spans
multiple decades.  As early as 1988, Carlson \& Ward examined the
potential use of natural language processing for the ``formulating
hypotheses'' task, a new item type under consideration for the GRE
test that would ask students to list all of the possible explanations
they could think of that would account for some observed phenomenon
(for example, a steady reduction in the mortality rate for a
particular population).  While this is a somewhat unique item type,
but it is quite similar in its fundamental scoring characteristics
(the fact that it is scored according to the correctness or semantic
appropriateness of a short, textual unit) to many other ``short
answer'' tasks that have been considered more recently.

Research on automated scoring of short-answer tasks continued at the
Educational Testing Service during the early 1990s
\cite{kaplan1991,kaplan1994,burstein1995}, and received broader
attention in the early 2000s, when a number of short answer scoring
systems were developed, including ETS' \emph{c-rater}
\cite{leacock03}, AutoMark \cite{mitchell02}, the Intelligent Essay
Assessor \cite{landauer03}, the Oxford-UCLES system
\cite{sukkarieh05}, and applications developed at the University of
Portsmouth \cite{callear01} and the University of Manchester
\cite{sargeant04}.  Some approaches to the task have relied heavily on
knowledge engineering, involving manual creation of patterns to
encapsulate correct answer types for particular questions
\cite{callear01,sukkarieh05}. Other approaches have aimed to use more
generic text similarity features to determine the distance between
students' responses and some ``gold standard'' answer or answers
\cite{landauer03,perez05,mohler11,meurers11,hahn12}.  Hybrid systems
have also been developed, in which some human involvement is required
for task-specific pattern creation or annotation, but other components
of the system use automatically-constructed features and statistical
calibration \cite{mitchell02,leacock03,nielsen08}.  There has been a
shift over time towards more fully-automated and statistically-based
systems, and away from those relying on manual knowledge engineering,
but the selection of methodology also depends on the exact type of
short answer questions targeted by each system.  For instance, the
tasks addressed by \newcite{mitchell02} required answers to include
specific well-defined concepts (see Figure \ref{mitchell-item-fig}),
and were therefore more amenable to a knowledge engineering approach,
whereas those addressed by Foltz et al. \shortcite{foltz2011} elicited
longer, less-constrained responses (see Figure \ref{foltz-item-fig}),
and were scored according to the evidence students gave of their
``depth of knowledge'', rather than for specific, correct concepts.

\begin{figure}[htb]
  \framebox{
    \parbox{\columnwidth} {
      \textbf{Question:}
      Why are some wildflowers highly scented with brightly colored petals?\\[5pt]
      \textbf{Rubric Answer:}
      To attract insects
    }
  }
\caption{Sample item from the UK Science National Test, from Mitchell et al., 2002}
\label{mitchell-item-fig}
\end{figure}

\begin{figure}[htb]
  \framebox{
    \parbox{\columnwidth} {
      \textbf{Question:}
      Explain how cows, grasses and bacteria interact within an environment,  In your explanation, be sure to include
      \begin{itemize}
        \item The role of each of the organisms in the environment
        \item How the organisms depend on one another
      \end{itemize}
    }
  }
\caption{Sample science item from the Maryland state assessment, from Foltz et al., 2011}
\label{foltz-item-fig}
\end{figure}

Some of these systems have recently seen operational use for scoring
consequential tests.  Foltz \shortcite{foltz2010} reported that
Pearson's Intelligent Essay Assessor was being used to score science
questions on the Maryland State Assessment.  Leacock and Chodorow
\shortcite{leacock03} also cite the use of ETS' \emph{c-rater} in a
state assessment context.  More opportunities for the use of such
systems in consequential testing systems are likely to emerge in
coming years, as well, as more state tests move from paper-and-pencil
administration to online formats, and as new multi-state tests are
developed.  Two state consortia (known as
PARCC\footnote{\texttt{http://smarterbalanced.org/}} and Smarter
Balanced\footnote{\texttt{http://parcconline.org/}}) have received funding
from the US Department of Education to develop next-generation tests
that can be used in multiple states, and incorporate innovative
technology to address a new set of standards for what children at
different grades should know and be able to do (the Common Core State
Standards\footnote{\texttt{http://corestandards.org/}}).  These tests are
slated to be launched in the 2014-2015 school year, and have
explicitly included the automated scoring of open-ended tasks as one
of their design desiderata.

Partly as a result of this increased commercial interest in the
automated scoring of short-answer questions, recent efforts have
arisen to empirically assess the state of the art in this field, and
to compare the performance of available systems.  One of these is the
``Joint Student Response Analysis and 8th Recognizing Textual
Entailment Challenge'' at SEMEVAL-2013, which added a new corpus of
student answers from the tutorial dialog context to the set of textual
entailment evaluation corpora \cite{dzikovska2012}.

The second organized effort to compare methods for short-answer
scoring is the The Automated Student Assessment Prize, the outcomes of
which directly motivate the current study, and from which the data
used here are drawn.

\subsection{The Automated Student Assessment Prize}
\label{sec:asap-description}
The Automated Student Assessment Prize (ASAP) is an effort funded by
the Hewlett Foundation, and organized by Open Educational Solutions,
to assess the current state of capabilities for automated scoring of a
wide range of open-ended student response tasks.  The short-answer
scoring competition conducted in 2012 is the second in a series of
such evaluations, with the first phase having focused on essay-length
responses \cite{shermis2012}, and future phases under discussion.  For
both phases of the ASAP competition, states participating in the
multi-state assessment consortia mentioned above have cooperated by
providing student responses to use as data.  The set of responses
these states have shared through the ASAP competition is the largest
publicly-available dataset for developing and evaluating automated
methods for scoring of short-answer questions to date.  

The organizers elected to host the competition through 
Kaggle\footnote{\texttt{http://kaggle.com}}, an online
platform for posting and running competitions (which typically involve 
computational modeling).  The Hewlett Foundation provided a
\$100,000 prize fund for the competition, and the contest rules
required that participants release all model submissions under
open-source licenses in order to be eligible for the cash prizes.  The
contest training data was released in late June, 2012, and the 256
participating teams had approximately 9 weeks in which to develop their
scoring methods.  Once the test data was released (without human
scores), participants had approximately one week to compute and submit
their final score predictions.  All score predictions were required to
be rounded to integers.  Most individuals or teams participating in
the competition did so under pseudonyms, so it is not entirely clear
what areas of expertise might have been reflected in the field.
However, since the tasks hosted on Kaggle typically do not involve
text processing, it is likely that many participants had more
experience with machine learning and applied statistics training than
with natural language processing.

The final rankings of prize-winning teams participating in the ASAP
short-answer scoring challenge are listed in Table
\ref{asap-results-table}, ranked by final weighted kappa on the test
set averaged across questions.  The top prize winner, Luis Tandalla, invested a great deal of
manual effort in developing regular expressions to cover many key
concepts that were referenced in student responses, and then trained a
top-level regression model to score responses based on the presence of
these concepts and the predictions of base models using lexical
features only \cite{tandalla2012}.  The remaining four prize winners
developed models that were quite similar in structure to one another,
also using a stacked prediction architecture.  Each of these models
used low-level classifiers to model the scores of responses based on
lexical features (the presence of particular words, stems, ngrams, or
character sequences), and the predictions of these classifiers were
combined using one or more higher-level models
\cite{zbontar2012,conort2012,jesensky2012,peters2012}.  Some models
incorporated other features in addition to the standard base learners
using lexical information, as indicated in Table \ref{asap-results-table}.

\begin{table}[tbh]
\begin{tabular}{@{}l@{\,}|@{\,}c@{\,}|@{\,}p{3.8cm}@{}}
\textbf{Team} & \textbf{Aggregate} & \textbf{Features} \\ 
\textbf{}     & \textbf{Weighted}  & \textbf{of Base} \\
\textbf{}     & \textbf{Kappa}     & \textbf{Learners} \\
\hline
L. Tandalla & 0.74717 & \parbox{3.8cm}{\vspace{.5ex} Lexical features\\ Task-specific regexes} \\
\hline
J. Zbontar & 0.73892 & \parbox{3.8cm}{\vspace{.5ex} Lexical features\\ Reduced-dimensionality lexical features} \\
\hline
X. Conort & 0.73662 & \parbox{3.8cm}{\vspace{.5ex} Lexical features\\ Reduced-dimensionality lexical features\\ Word frequency\\ Text well-formedness\\ Response length} \\
\hline
J. Jesensky & 0.73392 & \parbox{3.8cm}{\vspace{.5ex} Lexical features\\ Response length} \\
\hline
\parbox{2.1cm}{\vspace{.5ex} J. Peters \\ P. Jankiewicz} & 0.73094 & \parbox{3.8cm}{\vspace{.5ex} Lexical features\\ Reduced-dimensionality lexical features\\ Response length}
\end{tabular}
\caption{Prize-winning systems from ASAP short-answer scoring competition}
\label{asap-results-table}
\end{table}

These results demonstrate a number of points about the short answer scoring task, at least as presented in ASAP:
\begin{enumerate}
  \item Lexical features---which specific words and word sequences are used in a response---are strongly predictive of human scores, and can predict these scores even without other features.
  \item A predictive model architecture using stacked classifiers \cite{wolpert1992} seems to work well for this task---perhaps because providing different ``views'' of the feature set helps to address the sparsity of lexical features.
  \item Model performance can be improved somewhat by investing a great deal of effort to model specific response patterns (with regular expressions, for example), although such an approach would not scale well to large numbers of questions.
\end{enumerate}

In sum, the results of the ASAP competition suggest that relatively
superficial methods of automated short response scoring (leveraging
lexical features alone), can do quite well, and that while it is
possible to improve somewhat on such models, the methods needed to do
so are not scalable in any case.  The purpose of this paper is to more
closely examine the potential contribution of \emph{syntactically
  informed} features, that attempt to model response patterns above
the lexical level, and to investigate whether such features can be
developed in a generic way, so that manual work is not required to
support scoring responses to each new question.

\section{Data}
The data set used in this study was the same one used in the ASAP
short answer scoring challenge described above.  The questions and
student responses in this data set were contributed by multiple state
education departments, although the names of the participating states
were not provided by the organizers of the competition.  The
organizers did report that students' responses to some questions were
entered on a computer, while responses to other questions were
provided in handwritten form (and later transcribed).

The data set contains ten different short-answer questions, differing
in characteristics such as subject area, average response length, and
scoring scale, as shown in Table \ref{data-overview-table}.  All of
the ASAP questions were administered at Grade 10 in the US, except for
question 10, which was administered in Grade 8.  For each question, a
rubric is also provided, which outlines the criteria for assigning
scores to responses.\footnote{More detailed information about the ASAP
  questions and scoring guidelines is posted on the Kaggle site for
  the competition: \texttt{http://kaggle.com/c/asap-sas/data/}.

  The ASAP short answer data set is an important new resource for
  researchers working on analyzing short-answer questions responses.
  However, a few points are worth mentioning as context for the
  results of any studies using the corpus.  First, the data capture
  process for one question resulted in an artifact into the responses
  that cannot be straightforwardly reversed.  Space characters have
  been inserted, seemingly at random, into the responses to question
  10, often creating artificial breaks within words.  Second, a small
  number of responses seem to have been truncated, leaving only a
  short initial substring in place of the complete student response.
  (This speculation is based upon the observation that certain very
  short, seemingly incomplete responses received very high scores from
  both raters.)  Finally, note that the human agreement reported
  here is substantially higher for some tasks than has been observed
  in previously-reported studies using short-answer questions.  More
  information about states' scoring practices would help to clarify
  whether the two human ratings of these responses were truly provided
  independently of one another in all cases.  } The organizers report
that each response was scored by two independent human raters, and
Table \ref{data-overview-table} also reports the agreement between
these raters (on the training set) using the quadratic weighted kappa
measure widely used in studies of open-ended response scoring
(cf. \newcite{attali2010}; \newcite{williamson2012}).

\begin{table*}[tbh]
\begin{tabular}{r|c|c|c|c|c|c|c}
\textbf{Question No.} & \textbf{Subject} & \multicolumn{3}{|c|}{\textbf{No. Responses}}          & \textbf{Ave. Wds./Response} & \textbf{Score} & \textbf{Human Agmt.} \\ 
\textbf{}             & \textbf{Area}    & \textbf{Train} & \textbf{Leaderboard} & \textbf{Test} & \textbf{(Train Set)}        & \textbf{Scale} & \textbf{(Weighted $\kappa$)} \\ 
\hline
1 & Science & 1672 & 557 & 558 & 47.1 & 0--3 & 0.939 \\
\hline
2 & Science & 1278 & 426 & 426 & 59.2 & 0--3 & 0.922 \\
\hline
3 & ELA\footnote{English Language Arts} & 1808 & 406 & 318 & 47.7 & 0--2 & 0.746 \\
\hline
4 & ELA     & 1657 & 295 & 250 & 40.2 & 0--2 & 0.738 \\
\hline
5 & Biology & 1795 & 598 & 599 & 25.1 & 0--3 & 0.950 \\
\hline
6 & Biology & 1797 & 599 & 599 & 23.4 & 0--3 & 0.961 \\
\hline
7 & ELA     & 1799 & 599 & 601 & 41.1 & 0--2 & 0.970 \\
\hline
8 & ELA     & 1799 & 599 & 601 & 53.0 & 0--2 & 0.858 \\
\hline
9 & ELA     & 1798 & 599 & 600 & 49.7 & 0--2 & 0.810 \\
\hline
10& Science & 1640 & 546 & 548 & 41.4 & 0--2 & 0.883
\end{tabular}
\caption{Overview of questions and responses in the ASAP short answer scoring data set}
\label{data-overview-table}
\end{table*}

The responses to each ASAP task were divided by the organizers into
three partitions.  The \emph{training} partition contains
approximately 60\% of the responses to each question, and was intended
to be used for direct parametrization of automated scoring models
developed for the challenge.  Participants were provided with both
sets of human scores to all responses in the training partition.  The
\emph{leaderboard} partition contains another 20\% of the responses.
The texts of the leaderboard responses were available to participants
throughout the challenge, but the human scores for these responses
were not.  However, participants could submit their score estimates
for leaderboard responses through a web interface (at most, twice per
day) to learn how well these estimates agreed with human scores on the
leaderboard set, and to be ranked on a publicly visible
``leaderboard'' on this basis.  Finally, the remaining 20\% of the
responses made up the \emph{test} partition, which was provided to
participants without human scores at the very end of the competition,
and which was used as the basis for final rankings of systems.

\section{Model}
%  5 base classifiers and ensemble
The features described in the following section were used to train a
variety of different regression models to estimate the score to be
assigned to each response: 
\begin{itemize}
  \item simple least-squares linear regression, 
  \item ridge regression, 
  \item support vector regression (with an RBF kernel),
  \item random forests, and
  \item gradient boosting.
\end{itemize}
The parameters used in model training were determined by
cross-validation within the training set.

This setup is consistent with the popular
modeling approach of stacking used in the ASAP competition, in which
individual predictive features used by the top-level model are
themselves the outputs of classification or regression models (as are
many of the features described in Section \ref{feature-sec}).  And as many
participants chose to do in the ASAP competition, we have also
included a model variant which uses an ensemble as the top-level
model, in which the unrounded outputs of all five of the simple
regression models are averaged to produce the final ensemble
prediction.  For all models, rounding of scores is done as the final
step.

\section{Features}
\label{feature-sec}

\subsection{BASE Features}
\label{BASE-sec}
The four classes of features described in this section constitute the
``BASE'' set, which is intended to be representative of the features
commonly included in high-performing models from the ASAP challenge.

%%DCH6
%% Incomplete
\subsubsection{Bag of Words Features}
%% Bag of Words
The first set of base features included in our stacking classifier is
a feature that is itself the output of a model trained to predict
human scores based on the presence of specific words.  Such a
bag-of-words model using random forest regression was provided by the
competition organizers as a baseline for the ASAP challenge.

% TODO - CV training
% TODO - Binary representation

% TODO - How many BOW features?  What models (SVM, RF)? Feature selection/threshold?

%% Bag of Char Ngrams
Using the same methodology as the simple bag-of-words models, we also
included two features based on bags of character ngrams.  Character
ngrams tend to be useful for this task, because they capture lexical
information in a way that is insensitive to spelling errors.
% TODO - How many BOCN features?  What models (SVM, RF)? Feature selection/threshold?

Finally, we include two bag-of-stems features, trained using the same
method as above, but with all words pre-processed using the Porter
stemming algorithm.
% TODO - How many BOS features?  What models (SVM, RF)? Feature selection/threshold?

These three feature sets together will be referred to below as Bag of
Words (BOW) features.

\subsubsection{LDA Features}
For each question, two LDA topic
spaces were constructed using the MALLET toolkit \cite{mccallum2002}, one with 30
dimensions, and one with only ten.  The weight of each of these 40
topics for a given response was used as a feature.  These features are
meant to be roughly comparable to the reduced-dimensionality lexical
features used by ASAP participants.

\subsubsection{Well-formedness Features}
Five features were used to represent the degree to which each response
consisted of well-edited, grammatical English text.  These features
were based on the count of errors identified in four different
categories, as identified automatically by a system for grammatical
error detection \cite{attali06}, as well as a feature representing the sum
across error categories.

\subsubsection{Length Features}
Three features were used to represent length: the number of
characters, words, and sentences in each response.

\subsection{Syntactically-informed Features}
The four additional categories of features described in this section
are more ``syntactically-informed'' than those in the BASE set,
because they encode information about word sequences, syntactic units
or discourse units.

\subsubsection{Ngram Features}
This feature class included three ``bag-of-ngram'' features analogous
to the bag-of-words features described above.  Each included the top
1000 most frequent unagrams, bigrams and trigrams in a regression
model trained to predict the human score.  The three feature variants
used random forests, support vector machines, and ridge regression.

\subsubsection{Language Model Features}
Three features were included based on language models built using the
IRST Language Modeling Toolkit \cite{federico2008}.  Language models were trained on
responses to each question that received the highest score, one of the
two highest scores, and the score of zero. The perplexity of a
response with respect to each of these models was used as a feature.  

\subsubsection{Dependency Features}
This feature class included six ``bag-of-dependency'' features.  Each
included the top 1000 most frequent dependency triples (and possibly
other items; see below) extracted from responses in the training set
using the Stanford parser \cite{klein2003}  in a regression model trained to
predict the human score.  The feature variants included models
incorporating dependencies only, dependencies and single words, and
dependencies combined with ``partial'' dependencies (a lexical head
associated with a dependency relation, but not the other head with
which it is associated).  Each of the three variants was implemented
with both random forests and support vector machines, yielding a total
of six features. 

\subsubsection{$k$-Nearest Neighbors features}
Two features were developed that relate to the similarity of word
sequences used to respond to a question by different examinees.  To
represent the similarity between responses, we used a symmetric
version of the BLEU score
% ($\frac{1}{2}\mathrm{BLEU}(x,y)+\mathrm{BLEU}(y,x)$).  
$\left( \frac{\mathrm{BLEU}(x,y)+\mathrm{BLEU}(y,x)}{2} \right)$.  
We then
defined two features based on the set of eight training responses most
similar to a given response: the average score of these eight nearest
neighbors, and the distance-weighted average score.

\subsubsection{Discourse Segment Features}
This final feature class included five features based on a system for
segmenting short-answers into meaningful sub-units based on regular
expressions.  This model parses out sentences or clauses set off using
bullets, numbering, or discourse connectives such as
\emph{furthermore} or \emph{however}.  The features used are the count
of the total number of discourse units identified, the length of any
numbered list identified, the highest number identified associated
with a numbered bullet, and the number of discourse units headed by
discourse markers in two categories (the \emph{finally} category
indicating a conclusion, and the \emph{furthermore} category
indicating supplemental information).

\section{Experiments}
First, in order to demonstrate the strong performance of the ensemble
model on this data set relative to single top-level regression models,
we trained each model on the BASE feature set (cf. Section
\ref{BASE-sec}).  As Table \ref{eval-models-table} indicates, the
ensemble of regressors outperforms any individual model on two of the
three metrics, and is barely edged by gradient boosting on to the
third metric.

The results reported in Table
\ref{eval-models-table} are aggregated across all ten ASAP
questions, using three different metrics.  The average correlation of
\emph{unrounded} predicted scores with human scores provides the most
precise measure of model performance, since information is lost in the
rounding process, and in many testing contexts an unrounded item score
could be used directly to construct total test scores.  The average
quadratic weighted kappa, computed using rounded predictions, is
provided for comparison.  Finally, we also report the ``official''
metric used in the ASAP short-answer scoring competition, in which
weighted kappas for each task are manipulated with the Fisher
transform before averaging:

\begin{eqnarray}
  z & = & \frac{\sum_{i=1\ldots10} \frac{1}{2}\ln\frac{1+\kappa_i}{1-\kappa_i}}{10} \nonumber \\
  \kappa_{aggregate} & = & \frac{e^{2z}-1}{e^{2z}+1} \nonumber
\end{eqnarray}

\begin{table}[tbh]
\begin{tabular}{@{}r@{\,}|@{\,}c@{\,}|@{\,}c@{\,}|@{\,}c@{}}
\textbf{Model Type} & \textbf{Average}     & \textbf{Average}  & \textbf{ASAP} \\
\textbf{}            & \textbf{Correl.} & \textbf{Weighted} & \textbf{Weighted} \\
\textbf{}            & \textbf{}            & \textbf{Kappa}    & \textbf{Kappa} \\
\hline
linear regression & 0.775 & 0.740 & 0.749 \\
ridge regression & 0.785 & 0.746 & 0.755 \\
RBF-SV regression & 0.756 & 0.719 & 0.726 \\
random forest & 0.779 & 0.739 & 0.746 \\
gradient boosting & 0.785 & 0.751 & \textbf{0.761} \\
ensemble & \textbf{0.789} & \textbf{0.752} & 0.759
\end{tabular}
\caption{Results of each model type using BASE features (aggregated across ASAP questions)}
\label{eval-models-table}
\end{table}

In order to determine the effect of different feature categories on
overall performance, we then compared variants of the ensemble model
using the BASE feature set with variants that employed more features,
individually adding in each of the other feature categories described
above.  These results, as well as the performance results for a model
including ALL available features, are indicated in Table
\ref{eval-ablation-table}.

\begin{table}[tbh]
\begin{tabular}{@{}r@{\,}|@{\,}c@{\,}|@{\,}c@{\,}|@{\,}c@{}}
\textbf{Feature Set} & \textbf{Average}     & \textbf{Average}  & \textbf{ASAP} \\
\textbf{}            & \textbf{Correl.} & \textbf{Weighted} & \textbf{Weighted} \\
\textbf{}            & \textbf{}            & \textbf{Kappa}    & \textbf{Kappa} \\
\hline
BOW Only & 0.787 & 0.745 & 0.753 \\
BASE & 0.789 & 0.752 & 0.759 \\
BASE+NGRAMS & 0.789 & 0.749 & 0.756 \\
BASE+KNN-BLEU & 0.790 & 0.752 & 0.759 \\
BASE+SEGMENTS & 0.790 & 0.753 & 0.761 \\
BASE+LM & 0.792 & 0.750 & 0.758 \\
BASE+DEPS & 0.793 & \textbf{0.758} & 0.767 \\
ALL & \textbf{0.795} & \textbf{0.758} & \textbf{0.768} 
\end{tabular}
\caption{Results of ensemble model using each feature set (aggregated across ASAP questions)}
\label{eval-ablation-table}
\end{table}

As Table \ref{eval-ablation-table} shows, each of the new features
added to the BASE set adds some incremental predictive power to the
model, except for the bag-of-ngrams features, which slightly degrade
performance vs. the baseline.  The greatest increase in performance
for a single feature class comes from the inclusion of
dependency-based features, which yields a modest increase in all three
measures.  The addition of all five features to the BASE feature set
increases the overall weighted kappa using the ASAP contest metric by
approximately 0.009 to 0.768.  This increase may be small, but it is
not negligible, considering the magnitude of differences between
top-ranked participants.  (The difference between places 1 and 10 in the final
ranking of participants on the public leaderboard was only 0.011.)
% Note that of the ten questions in the ASAP data set, the
% ALL-features model demonstrated improved performance over the BASE
% model on seven.
Note that the BASE feature set performs slightly better than a
model using only bag-of-words features (including features based on
stems and character ngrams).

%% Paired t tests seem to show significant difference only between BOW and top-performing features
%% Nothing significantly better than BASE (ALL > Base at p=.067)

\begin{figure*}
  \begin{center}
    \includegraphics[width=\textwidth]{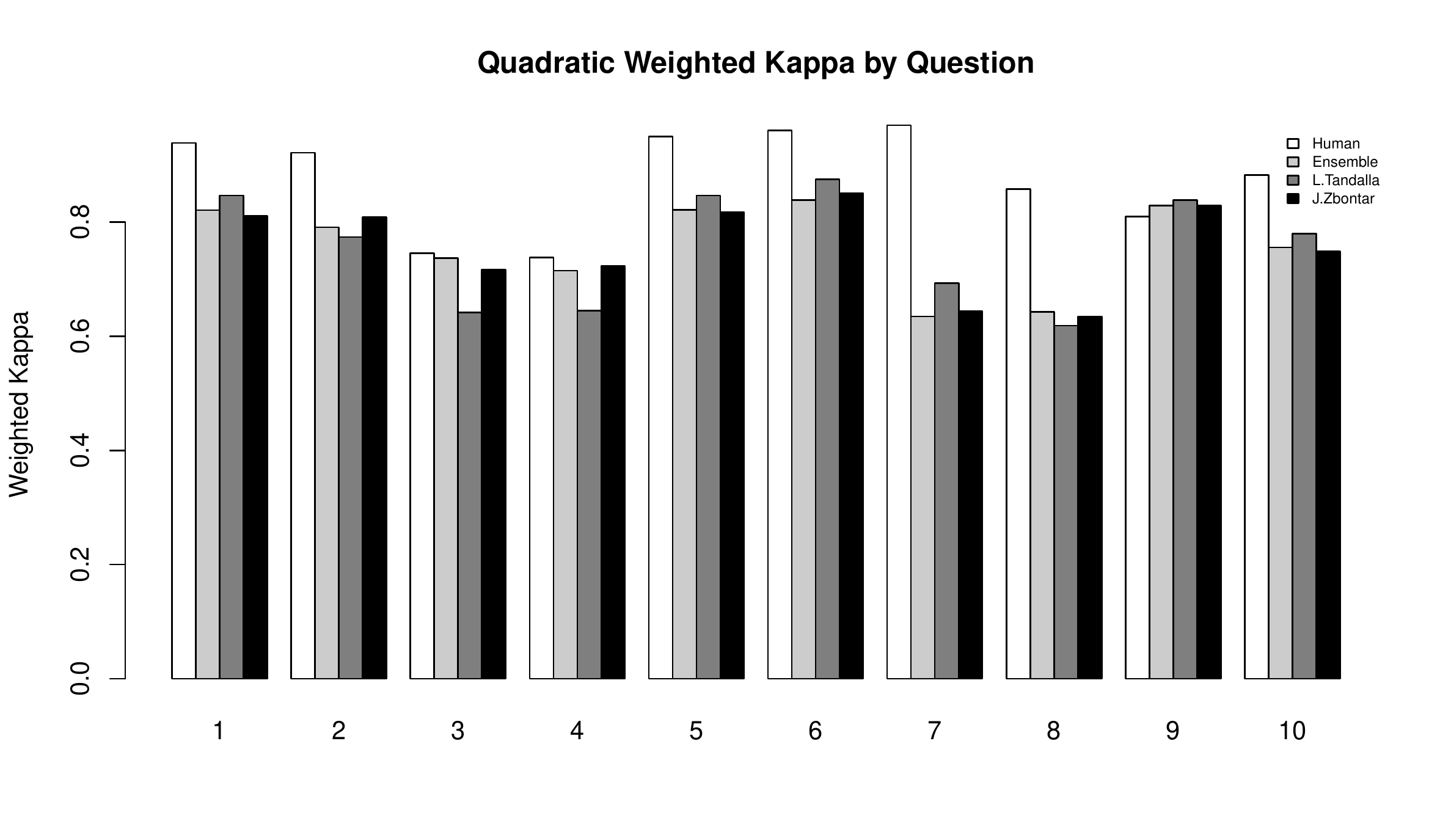}
  \end{center}
\caption{Inter-rater agreement for human scores, ensemble model (ALL features), and two top models from ASAP challenge, across questions}
\label{eval-tasks-figure}
\end{figure*}

Figure \ref{eval-tasks-figure} shows the performance of the ensemble
model across all ten ASAP questions, as well as the cross-task
performance of two top-ranked systems from the
competition\footnote{Since participants made their code available as
  open source, we were able to reproduce their models and generate
  more detailed statistical evaluations.}.  The agreement of predicted
scores and human scores varies both according to the reliability of
human scoring (cf. Table \ref{data-overview-table}) and according to
the difficulty of modeling the distinctions made in the scoring
rubrics for each question.  For example, questions 7 and 8 call for
students to draw conclusions based on a reading passage, and to
support those conclusions with examples.  In part because of the
supra-lexical semantic requirements this imposes on responses, the
reliability of automated scoring lags behind that of human scoring for
these items.  By contrast, questions 5 and 6 ask students to list
specific biological concepts or processes, and for these items
automated scoring performance is higher both in absolute terms and
relative to human scoring.

%% \begin{table}[tbh]
%% \begin{tabular}{r|c|c}
%% \textbf{Question No.} & \textbf{Correlation}     & \textbf{Weighted} \\
%% \textbf{}             & \textbf{}                & \textbf{Kappa} \\
%% \hline
%% 1 & 0.846 & 0.821 \\
%% 2 & 0.820 & 0.791 \\
%% 3 & 0.766 & 0.737 \\
%% 4 & 0.781 & 0.715 \\
%% 5 & 0.856 & 0.822 \\
%% 6 & 0.864 & 0.839 \\
%% 7 & 0.688 & 0.635 \\
%% 8 & 0.676 & 0.643 \\
%% 9 & 0.859 & 0.829 \\
%% 10 & 0.793 & 0.756
%% \end{tabular}
%% \caption{Results for ensemble model (ALL features) on each ASAP question}
%% \label{eval-tasks-table}
%% \end{table}

\section{Conclusions and Discussion}
\subsection{Performance relative to ASAP systems}
The first point to note in connection with the results reported above
is that the final ensemble model's prediction accuracy on the ASAP
leaderboard data is lower than the highest scores posted during the
competition itself.  The top final weighted kappa score on the ASAP
public leaderboard was 0.772, compared to our result of 0.768.  There
are a number of reasons for this.  

Firstly, stacking models such as our ensemble model and the models
applied by the top-ranked ASAP participants have a number of free
parameters that can be iteratively manipulated in order to improve the
models' performance.  Learning parameters of the base-level and
top-level regression models can be tweaked, features can be added or
removed, and feature variants can be introduced in order to optimize
prediction results.  Since ASAP participants had the opportunity to
submit predictions to the Kaggle server twice a day for two months
(receiving performance measures on the leaderboard set each time),
there was ample opportunity to fit these parameters to the leaderboard
data.  In contrast, the models described here were not optimized in
any way using information from the leaderboard set.  While the ASAP
challenge also had an independent test set (the ``private
leaderboard''), the human scores on that set have not been released
publicly, so it could not be used for the current study.

The second reason why the ensemble model's performance trails that of
the top ASAP models is that it uses exactly the same features and
parameterization across all ten questions.  Because our aim was to
examine the contribution of different feature categories to overall
scoring accuracy, rather than to achieve the highest possible scoring
accuracy through model optimization, we did not optimize the
configuration of the model to each question individually (as most ASAP
participants did).

Finally, in contrast to some ASAP systems, our ensemble model does not
use any special method for rounding of scores; scores are simply
truncated to the allowable range, and rounded to the nearest integer.
As noted above, unrounded scores are more appropriate for most testing
purposes.

\subsection{Role of syntactically-informed features}
The question this paper set out to address was to reexamine the
seeming lesson of the ASAP short answer scoring challenge, that
lexical features alone could not be improved upon except by investing
considerable effort in task-specific knowledge engineering.  The
results presented in Table \ref{eval-ablation-table} above indicate
that more syntactically-informed features can, indeed, contribute to
improved short-answer scoring performance.  The use of all new
features together improves the aggregate weighted kappa by about
0.009, with the greatest increase coming from the incorporation of
syntactic dependency information, the feature class with the clearest
link to syntax. This is particularly remarkable given that many
student answers contain numerous misspellings and grammatical errors,
making them difficult to parse reliably.

The incremental predictive value of these features was modest, but
there may be motivation for including such features in short-answer
scoring systems beyond their empirical effects on scoring accuracy.
Since the scoring rubrics for these items claim to be sensitive to
characteristics of responses beyond simple word choice, including
higher-level features will improve the systems' \emph{validity}, the
degree to which it will actually measure the skills and knowledge that
items are supposed to test.  If students are not able to get a high
score simply by producing a response that includes relevant
vocabulary, this will reduce the risk of \emph{negative washback}, or
unproductive learning strategies directed solely toward optimizing
test performance.

\subsection{Future work}
There are a number of areas in which this work could be extended.
First, of course, the use of syntactic dependencies does not come
close to exhausting the space of feature types that more directly
reflect the syntactico-semantic relationships that short answers are
supposed to encode.  Features based on semantic role labeling and
paraphrase detection, for example, may offer additional benefits.
There may also be some gains to be had from tailoring features to a
particular content area (for example, Biology or reading
comprehension).

Another important question to address is how the particular setting of
the ASAP task affects the results achieved.  Parameters such as
training sample size, the number of tasks to be modeled concurrently,
and the content areas to be scored, could significantly influence
the findings.

%% \section*{Acknowledgments}
%% \DCH{TODO}

\nocite{carlson1988}

\bibliographystyle{naaclhlt2013}
\bibliography{henry}

\end{document}